\documentclass[sigconf]{aamas} 

\usepackage{balance} % for balancing columns on the final page

\usepackage{algorithm}
\usepackage{booktabs}
\usepackage{amsmath}
\usepackage{xcolor}
\usepackage{enumitem}

\usepackage[utf8]{inputenc} % allow utf-8 input
\usepackage[T1]{fontenc}    % use 8-bit T1 fonts
\usepackage{hyperref}       % hyperlinks
\usepackage{url}            % simple URL typesetting
\usepackage{booktabs}       % professional-quality tables

\usepackage{amsfonts}       % blackboard math symbols
\usepackage{nicefrac}       % compact symbols for 1/2, etc.
\usepackage{microtype}      % microtypography
\usepackage{xcolor}         % colors

\usepackage{comment}
\usepackage{subcaption}
\usepackage{macros}
\usepackage{soul}
\sethlcolor{yellow}
\usepackage{array}

\DeclareRobustCommand{\osher}[1]{{\color{green}{Osher: #1}}}

\definecolor{goalAdapt_color}{HTML}{6121ef}

\definecolor{domainAdapt_color}{HTML}{f7934b}

\definecolor{inference_color}{HTML}{3ec163}

\definecolor{customgray}{rgb}{0.65, 0.65, 0.65} % Adjust the RGB values (0 to 1)

\newtheorem{definition}{Definition}

\usepackage{newfloat}
\usepackage{listings}
\DeclareCaptionStyle{ruled}{labelfont=normalfont,labelsep=colon,strut=off} % DO NOT CHANGE THIS
\floatstyle{ruled}
\newfloat{listing}{tb}{lst}{}
\floatname{listing}{Listing}

\setcopyright{ifaamas}
\acmConference[AAMAS '26]{Proc.\@ of the 25th International Conference
on Autonomous Agents and Multiagent Systems (AAMAS 2026)}{May 25 -- 29, 2026}
{Paphos, Cyprus}{C.~Amato, L.~Dennis, V.~Mascardi, J.~Thangarajah (eds.)}
\copyrightyear{2026}
\acmYear{2026}
\acmDOI{}
\acmPrice{}
\acmISBN{}

\author{Osher Elhadad}
\affiliation{
  \institution{Department of Computer Science, Bar Ilan University}
  \city{Ramat Gan}
  \country{Israel}}
\email{oshereli2@gmail.com}

\author{Felipe Meneguzzi}
\affiliation{
  \institution{Department of Computer Science, Aberdeen University}
  \city{Aberdeen}
  \country{Scotland, UK}}
\email{felipe.meneguzzi@abdn.ac.uk}

\author{Reuth Mirsky}
\affiliation{
  \institution{Department of Computer Science, Tufts University}
  \city{Medford}
  \country{MA, USA}}
\email{reuth.mirsky@tufts.edu}

\title{GRAIL: Goal Recognition Alignment through Imitation Learning}

\begin{abstract}
Understanding an agent's goals from its behavior is fundamental to aligning AI systems with human intentions. 
Existing goal recognition methods typically rely on an optimal goal-oriented policy representation, which may differ from the actor's true behavior and hinder the accurate recognition of their goal. %This limits these robustness in real-world, human-in-the-loop settings.
To address this gap, this paper introduces Goal Recognition Alignment through Imitation Learning (GRAIL), which leverages imitation learning and inverse reinforcement learning to learn one goal-directed policy for each candidate goal directly from (potentially suboptimal) demonstration trajectories. 
By scoring an observed partial trajectory with each learned goal-directed policy in a single forward pass, GRAIL retains the one-shot inference capability of classical goal recognition while leveraging learned policies that can capture suboptimal and systematically biased behavior.
Across the evaluated domains, GRAIL increases the F$_1$-score by more than 0.5 under systematically biased optimal behavior, achieves gains of approximately 0.1–0.3 under suboptimal behavior, and yields improvements of up to 0.4 under noisy optimal trajectories, while remaining competitive in fully optimal settings.
This work contributes toward scalable and robust models for interpreting agent goals in uncertain environments.
\end{abstract}


\newcommand{\BibTeX}{\rm B\kern-.05em{\sc i\kern-.025em b}\kern-.08em\TeX}

\begin{document}

%%% The following commands remove the headers in your paper. For final 
%%% papers, these will be inserted during the pagination process.

\pagestyle{fancy}
\fancyhead{}

%%% The next command prints the information defined in the preamble.

\maketitle 

%%%%%%%%%%%%%%%%%%%%%%%%%%%%%%%%%%%%%%%%%%%%%%%%%%%%%%%%%%%%%%%%%%%%%%%%

\section{Introduction}
Goal Recognition (GR) aims to infer an agent's underlying objectives from its observable behavior \cite{sukthankar2014plan, mirsky2021introduction}. 
Accurate and timely goal inference is critical for interactive and cooperative systems. 
Such systems include human-robot interaction where aligning with human intent ensures safe and supportive assistance~\citep{massardi2020parc,trick2019multimodal,scassellati2002theory}
as well as multiagent scenarios, where predicting others' goals facilitates adaptive coordination and strategic reasoning~\citep{rabkina2019analogical,kaminka2001new,sukthankar2011activity,bansal2019beyond}.

By recognizing latent goals in real-time, AI systems can anticipate needs~\cite{shvo2020active}, proactively assist~\cite{oh2014probabilistic}, and detect anomalies, model discrepancies~\cite{sreedharan2018handling}, or misalignments~\cite{bernstein2002complexity, masters2019goal, zhuang2020consequences, mechergui2024goal}.
However, existing GR approaches often rest on a restrictive assumption that observed agents act (near-)optimally with respect to a predefined objective~\cite{masters2021s}. Classical plan recognition methods~\cite{ramirez2010probabilistic}, inverse planning~\cite{baker2007goal, baker2009action}, and more recent reinforcement learning (RL)-based goal recognizers~\cite{amado2022goal} %add if accepted: nageris2024goal
build models that expect rational or reward-maximizing behavior. 
This leads to brittle performance when applied to agents whose actions are noisy, constrained, or systematically biased. 
Human behavior, for example, frequently deviates from optimality in meaningful ways, reflecting cognitive limitations, preferences, social norms, or learned heuristics. 
As noted in~\citet{lindner2022humans}, such behaviors often cannot be captured by Boltzmann rationality models, and as shown in~\citet{masters2019goal}, accommodating subrational or irrational actors requires new modeling assumptions. Thus, existing GR systems often fail to align with real-world agents. In addition to these assumptions, practical deployment of GR further faces significant computational challenges. 
Many methods rely on RL to train a policy per candidate goal, often requiring millions of interactions with the environment. 
Additionally, at inference time, these methods may require planner invocations~\cite{ramirez2009plan} for each observed trajectory, creating latency bottlenecks in interactive or time-sensitive applications. Such constraints highlight a broader issue: real agents frequently exhibit patterned departures from optimality.

\begin{figure}
    \centering
    \includegraphics[trim=0 12cm 17.5cm 0, clip, width=0.9
\linewidth]{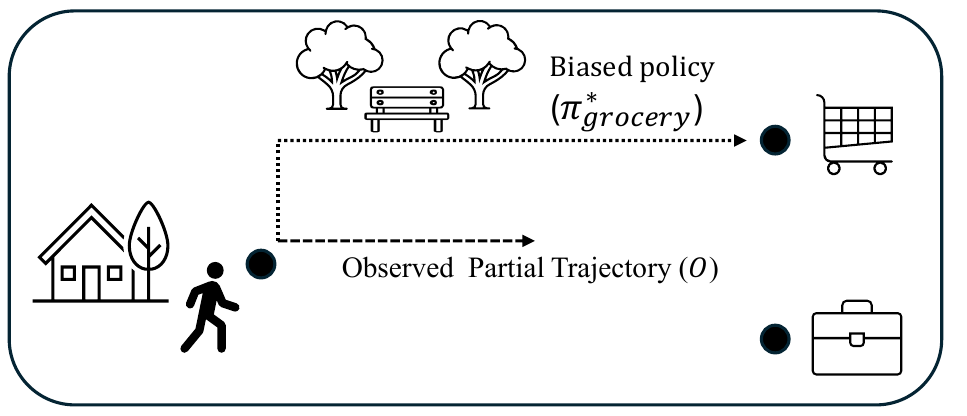}
    \caption{A navigation setup illustrating how ignoring systematic human biases like preference for park routes can lead to reduced inference accuracy in standard GR systems.}
    \label{fig:nav}
\end{figure}

Human navigation often deviates from shortest-path optimality, reflecting systematic preferences such as minimizing turns or following familiar landmarks~\cite{brown2021unified}. 
Consider the scenario in Figure~\ref{fig:nav} where a person leaves home with two potential destinations: a grocery store to the northeast and their workplace to the southeast. 
Although the two main routes to the grocery, going east then north or north then east, take the same time, the person consistently prefers the path that passes through a scenic park located to the north. 
When observing this person walking directly east, an optimality-based goal recognizer would assign similar likelihoods to the grocery and workplace, since both are consistently equidistant throughout the observation.
However, this overlooks the person's habit of going to the grocery via the park. 
In contrast, by learning goal-directed behavior from previous walks, the GR system can capture that this person rarely begins by walking east when going to the grocery. 
Therefore, it correctly infers that the workplace is a more likely goal, even though both destinations are equally optimal under standard assumptions.

To correctly recognize goals in such cases, and to address the additional challenges mentioned above, we develop \textbf{GRAIL (Goal Recognition Alignment through Imitation Learning)}, a novel framework that formulates GR as a collection of imitation learning (IL) problems, one for each candidate goal $g$.
% \correct{
% For each goal, GRAIL learns a policy $\pi_g$ using an IL technique that captures the statistical structure of the behavior under that goal: Behavioral Cloning~\cite{bain1995framework}, Generative Adversarial Imitation Learning  (GAIL)~\cite{ho2016generative}, and Adversarial Inverse Reinforcement Learning (AIRL)~\cite{fu2017learning}. 
% At test time, given a trajectory $\trajectory$, GRAIL computes a compatibility score $S_g(\trajectory)$ for each goal, reflecting how likely it is that an agent pursuing $\trajectory$ to be aiming to achieve $g$. Then, it selects the most likely goal, $\hat{g}$, using one-shot inference: $\hat{g} = \arg\max_g S_g(\trajectory)$.
% }{}{Too much technical detail for an intro. If you want to use this text later as intuition, fine, but let's not keep repeating ourselves.}

Unlike conventional GR methods that treat deviations from optimality as noise, our IL-based method is aligned with the realities of human and agent behavior, that can be (1) suboptimal or (2) systematically biased-optimal, thereby allowing GR systems to align with the actual behavior model of agents that are not idealized optimal agents. 
This novel GR formulation not only enhances robustness to real-world agent behavior but also enables fast, planner-free GR at test time, making GRAIL suitable for interactive scenarios where low inference-time latency is critical. 
We show that a variety of IL algorithms can be repurposed for one-shot goal inference without any planner or reward modeling at test time. This connects fields (GR, IRL, IL) that were previously treated as disjoint.  
%Experiments across MiniGrid and PandaGym domains demonstrate that GRAIL achieves substantial gains in both recognition accuracy and \hl{inference-time efficiency}, particularly in settings characterized by noise, preference-driven trajectories, or model mismatch. Our results highlight GRAIL as a scalable, alignment-aware solution for interpreting agent goals in \hl{controlled, uncertain, closed-set goal} environments.

Experiments across MiniGrid and PandaReach show that our imitation-based goal recognizers achieve substantial gains in recognition quality and practical efficiency.
In systematically biased optimal MiniGrid tasks, all GRAIL variants achieve near-perfect GR, whereas the Q-learning baseline remains at only 0.3–0.5. Under suboptimal behavior, GRAIL delivers consistent gains of roughly 0.1-0.3 F$_1$-score, particularly in the harder multi-goal settings. In PandaReach, our method improves F$_1$-score by up to about 0.4 over the actor–critic baseline in noisy optimal conditions, while matching or only slightly underperforming compared to RL-based goal recognition under clean optimal demonstrations.
These results indicate that GRAIL scales gracefully with noise, preference-driven trajectories, and model mismatch in controlled, closed-set goal environments, while preserving lightweight, forward-pass-only inference suitable for real-time use.

\section{Background}
\label{sec:background}
GRAIL is an interdiciplinary algorithm that solves a \textbf{Goal Recognition (GR)} problem using \textbf{Imitation Learning (IL)} and \textbf{Inverse Reinforcement Learning (IRL)} as components of its solution technique. To better place the contribution with respect to each of these domains, we refer the reader to Table \ref{tab:paradigm-comparison}. GR is a \emph{problem setting}: given a partial observation sequence and a set of candidate goals, infer which goal the observed agent is pursuing--essentially, goal classification at inference time.
In contrast, IL and IRL are \emph{learning paradigms}: IL learns a policy from demonstration trajectories to mimic expert behavior, while IRL recovers a reward function (and often a derived policy) that explains demonstrated behavior under known dynamics. Like IL/IRL, GR may use demonstrations at training time; however, at inference time GR takes a partial trajectory plus candidate goals as input and outputs a goal label, whereas IL/IRL produce policies or rewards. Next, we provide the required background and definitions from each of these paradigms.

\begin{table}[t]
\centering
\small
\begin{tabular}{|p{0.04\linewidth}|p{0.16\linewidth}|p{0.11\linewidth}|p{0.16\linewidth}|p{0.29\linewidth}|}
\hline
 & \textbf{Object of study} & \textbf{Typical output} & \textbf{Typical use} & \textbf{Role in this paper} \\
\hline
\textbf{GR} & Observed behavior \& candidate goals & Goal label or posterior & Recognize intentions & Core task we solve \\
\hline
\textbf{IL} & Expert demos & Policy $\pi$ & Learn behavior from examples & Learn goal-directed policy per goal \\
\hline
\textbf{IRL} & Expert demos \& known dynamics & Reward $R$ (+ policy) & Recover preferences for planning/RL & Via AIRL, obtain goal-directed policy head per goal; no reward-based planning at test \\
\hline
\end{tabular}
\caption{Role summary of GR, IL, and IRL. GR is a \emph{problem}; IL/IRL are \emph{learning paradigms} used here to build goal-directed policies for solving GR.}
\label{tab:paradigm-comparison}
\end{table}

We start by using the GR definition used in Amado et al.~\cite{amado2022goal}: %If accepted - add other surveys/book

\begin{definition}
\textbf{Goal Recognition (GR)} is the problem of inferring which objective an agent is pursuing from a partial sequence of observations. It is defined as a tuple $(\states,\actions,\goals,\observations)$, where: $\states$ is the state space, $\actions$ is the action space, $\goals = \{g_1,\dots,g_K\}$ is a finite set of candidate goals, and
 $\observations = (o_1,\dots,o_T)$ is the observation sequence (e.g., when observations are state-action pairs, $\observations = (\astate_0,\action_0,\dots,\astate_T)$) where $T$ is the number of observations in the sequence.
\end{definition}

Solving a GR problem requires selecting $g \in \goals$ that maximizes the likelihood of $\observations$.
This formulation captures the core setting of GR: a system must interpret observed actions and infer which of several possible goals best explains them. 
Most classical approaches treat this as an inverse-planning problem: they assume an optimal planner or a goal-directed RL policy and score each goal by how well its planner or policy could have produced the observations. 
% \correct{In contrast, GRAIL (Goal Recognition As Imitation Learning) replaces these restrictive assumptions with a purely data-driven paradigm. 
% For each candidate goal $g\in\goals$, we directly learn a goal-conditioned policy $\policy_g(a\mid s)$ from the corresponding demonstration set $\dems_g$.}{}{Not the place for this. You already made this contrast earlier in the intro. Focus on the background here.}
%

To position GR within the broader context of sequential decision-making, we briefly review reinforcement learning (RL), which formulates decision-making as trial-and-error interactions with an environment to maximize cumulative reward.

A \textbf{Markov decision process} (MDP) is a tuple $(\states, \actions, \transition, \reward, \discount)$, where: $\states$ is the state space; $\actions$ is the action space; $\transition(\astate' \mid \astate,\action)$ defines the transition dynamics; $\reward(\astate,\action)$ is the reward function; and $\discount \in [0,1)$ is the discount factor.

A solution to an RL problem is a policy $\policy(\action \mid \astate)$ that seeks to maximize expected discounted return:
$
\mathbb{E}_{\policy} \left[ \sum_{t=0}^\infty \gamma^t \reward(\astate_t, \action_t) \right].
$
\noindent
RL is well-suited for goal-directed behavior generation, but it assumes access to a reward function and often requires extensive interaction with the environment. 
Both assumptions are problematic for GR settings, where observed behavior is passive and the observed agent's rewards are unknown.

\textbf{Inverse Reinforcement Learning (IRL)} aims to recover a reward function that explains observed agent behavior~\cite{arora2021survey}. 
IRL algorithms use demonstrations to infer an agent's preferences through a model of their reward, typically requiring full trajectories and known dynamics.
In our work, we leverage IRL (specifically, AIRL) solely to obtain goal-directed policy heads.

\textbf{Imitation Learning (IL)}
comprises a family of techniques where an agent learns to perform tasks by mimicking expert behavior, typically using demonstrations of state-action trajectories. 
Unlike IRL, which seeks to infer the underlying reward function, IL focuses on directly learning a policy that reproduces expert-like behavior, often without explicitly modeling or recovering the reward structure driving the expert's decisions.
Given $N$ expert demonstrations $D = \{\trajectory_i\}_{i=1}^N$, where each trajectory $\trajectory_i$ is a sequence of states and actions $\trajectory_i = (\astate_0^i, \action_0^i, \dots)$, IL aims to learn a policy $\pi(\action \mid \astate)$ that minimizes divergence from the expert behavior distribution.
% \textcolor{blue}{In this paper, we use IL methods to learn one goal-directed policy per candidate goal, which we then employ to solve GR tasks.}
% \end{definition}

\section{Goal Recognition Alignment through Imitation Learning}
\label{sec:method}

\begin{figure}[t]
    \centering
    \includegraphics[trim=0 2cm 16cm 0, clip, width=1\linewidth]{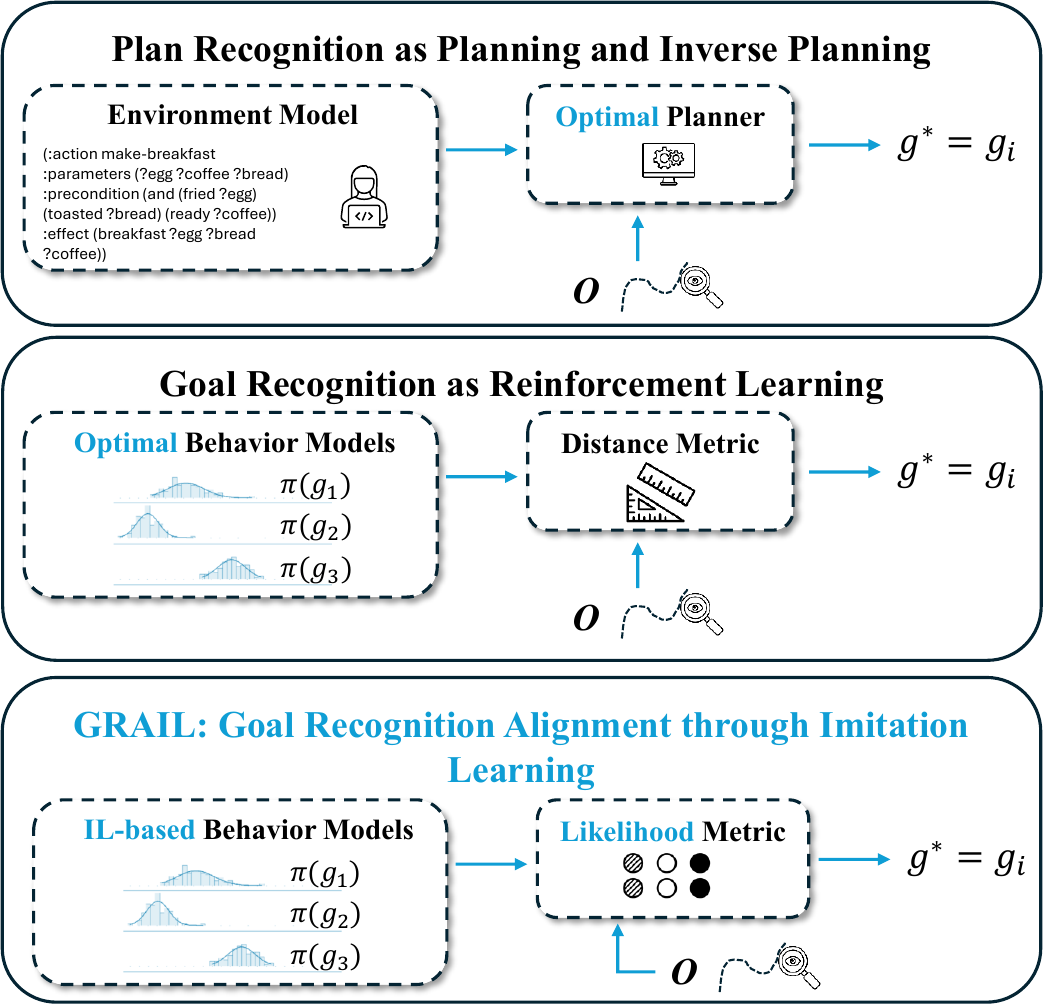}
    \caption{A comparison between model-based GR approaches (top), the RL-based GR approaches (middle), and our proposed framework (bottom).}
    \label{fig:comparison}
\end{figure}

We begin by formally defining the problem of GR from demonstrations. 
Similarly to Table~\ref{tab:paradigm-comparison}, which compares GR, IRL, and IL, Figure~\ref{fig:comparison} contrasts GR approaches and situates GRAIL within this space.

\begin{definition}[Goal Recognition from Demonstrations (GRD)]
A GRD task is specified by the tuple
$\bigl(\states,\actions,\goals,\dems_, \trajectory_\mathrm{obs}\bigr)$
where:
\begin{itemize}
  \item $\states$ is the set of all possible states;
  \item $\actions$ is the set of all actions;
  \item $\goals = \{g_1,\dots,g_K\}$ is a finite set of candidate goals;
  \item $\dems = \{\dems_{g}\}_{g\in\goals}$ is a set of demonstrations per goal. 
  \item $\trajectory_\mathrm{obs}=(\astate_0,\action_0,\dots,\astate_{T})$ is a new, partially observed trajectory. %\osher{I think we should change trajectory to observation sequence, and specifically here to $\observations = (\langle s_0,a_0 \rangle,\dots,\langle s_T, a_T \rangle)$ as in GR definition, what do you think?} 
\end{itemize}
Each $\dems_g = \{\trajectory^i_g\}_{i=1}^{N_g}$ is a set of trajectories $\trajectory^i_g = (\astate^i_0, \action^i_0, \dots, \astate^i_{T_i})$ generated by an agent pursuing goal $g$. 
This formulation does not require all trajectories for a goal $\{\trajectory^i_g\}_{i=1}^{N_g}$ to have a single starting point, nor a single end point, they should merely reflect the expected behavior of the goal-directed agent. 
The objective of a GRD task is to select the goal $\hat g\in\goals$ that best explains $\trajectory_\mathrm{obs}$.
\end{definition}

The key insight behind GRAIL is that whether behavior is optimal, systematically biased, or noisy, it still leaves a distinctive \emph{distributional fingerprint} on the trajectories.
IL methods are ideally suited to capture these complex fingerprints.  
Once we have trained each $\policy_g$, we perform one-shot inference by evaluating
\[
\states_g(\trajectory_\mathrm{obs})
\;=\;
\ell\bigl(\trajectory_\mathrm{obs};\pi_g\bigr),
\]
where $\ell$ is a goal-agnostic scoring function.  We can then infer $\inferredgoal$ by choosing the goal that maximizes this score.
\[
  \inferredgoal \;=\;\arg\max_{g\in\goals}\states_g(\trajectory_\mathrm{obs}).
\]

% \frm[inline]{Same as above, you don't need to narrate that you move to the next step (and why it's good). Just continue describing the backends.}
% We move on to detail the specific IL backends we employ, describe our choice of scoring functions for both discrete (MiniGrid) and continuous (PandaReach) tasks, and provide theoretical and empirical validation of why training on demonstrations yields reliable, fast, and robust goal inference under a wide range of real-world conditions.

% \begin{algorithm}[b]
% \caption{GRAIL Offline: Train per‐goal IL policies}
% \label{alg:grail-offline}
% \begin{algorithmic}[1]
%   \Require State space $\states$, action space $\actions$, goal set $\goals$, demonstrations $\{\dems_g\}$
%   \ForAll{$g\in\goals$}
%     \State $\pi_g \gets \mathrm{IL\_Learn}(\dems_g)$ \Comment{via BC, GAIL, or AIRL}
%   \EndFor
%   \State \Return $\{\pi_g\}_{g\in\goals}$
% \end{algorithmic}
% \end{algorithm}

% \begin{algorithm}[t]
% \caption{GRAIL Online: One‑Shot Inference}
% \label{alg:grail-online}
% \begin{algorithmic}[1]
%   \Require Policies $\{\pi_g\}$, trajectory $\trajectory=(s_0,a_0,\dots,s_T)$
%   \ForAll{$g\in\goals$}
%     \State $S_g \gets \mathrm{Score}(\pi_g,\trajectory)$
%   \EndFor
%   \State \Return $\displaystyle\arg\max_{g\in\goals} S_g$
% \end{algorithmic}
% \end{algorithm}

GRAIL alternates between two stages. 
First, using \emph{offline policy learning}, it trains a separate IL policy for each potential goal. 
Then, in an \emph{online, one-shot inference} phase, it scores a new observation sequence against every trained policy to pick the most plausible goal.

\subsection{Offline Policy Learning from Demonstrations}  

For each goal $g \in \goals$, we collect a dataset of demonstrations $\dems_g$ of an agent that aims to achieve $g$, and invoke our chosen IL routine to produce a goal-directed policy $\pi_g(\action \mid \astate)$.  
%In pseudocode this is exactly Algorithm \ref{alg:grail-offline}, where each $\mathrm{IL\_Learn}(\dems_g)$ call denotes one of these three algorithms. 
By the end of this phase we possess a ``bank'' of $K$ policies, each of which has directly absorbed the characteristic behavioral fingerprint of reaching its assigned goal, including any systematic biases or suboptimal quirks present in the demonstrations. 
In this paper, we explore three common IL algorithms at the backbone for GRAIL:
\begin{itemize}
    \item BC: Behavioral Cloning~\cite{bain1995framework}:  
  Treats imitation as supervised learning over state-action pairs. The goal-specific policy is trained by maximizing the likelihood of expert actions.
  % \[
  % \min_{\pi_g} \;\E_{(\state,\action)\sim \dems_g} \left[ -\log \pi_g(\action \mid \state) \right]
  % \]

  \item GAIL: Generative Adversarial Imitation Learning~\cite{ho2016generative}:  
  Uses adversarial training to match the occupancy distributions of expert and policy rollouts. 
  A discriminator %\(dems_g(\state,\action)\) 
  distinguishes expert versus learner trajectories.
  % :
  % \begin{align*}
  % \min_{\pi_g} \;\max_{\dems_g} \;&\; \E_{(\state,\action)\sim \dems_g} \left[ \log \dems_g(\state,\action) \right] \\
  % &+ \E_{(\state,\action)\sim \pi_g} \left[ \log(1 - \dems_g(\state,\action)) \right]
  % \end{align*}
  \item  AIRL: Adversarial IRL~\cite{fu2017learning}:  
  Learns both a reward function \(\reward_g(s,a)\) and a policy \(\pi_g\) such that the induced behavior explains expert data. The reward serves as a latent rationalization of actions.
% :
%   \[
%   \dems_g(\state,\action) = \frac{\exp(\reward_g(\state,\action))}{\exp(\reward_g(\state,\action)) + \pi_g(\action \mid \state)}
%   \]

\end{itemize}

\begin{table*}[tb]
\centering
\renewcommand{\arraystretch}{1.2}
\begin{tabular}{m{2cm} m{2cm} m{9cm}}
    \toprule
    \textbf{Figure} & \textbf{Domain} & \textbf{Details} \\
    \midrule
    \includegraphics[width=1.5cm]{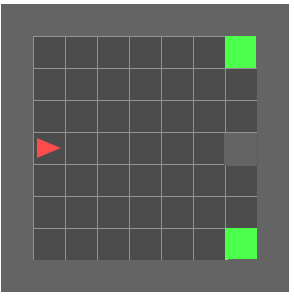} 
    & \raggedright\arraybackslash \textbf{MiniGrid} 
    & \raggedright\arraybackslash \textit{Environments}: MiniGrid-9x9 environment with an obstacle at 7x4. \newline \textit{Motivation}: Discrete navigation; comparison with GR baselines; interpretable and controllable (e.g. easy for designing suboptimal behavior). \newline
      \textit{States}: Position (x,y) + direction of agent - one of four compass directions (Discrete). \newline \textit{Actions}: Turn, move forward, stay in place (Discrete). \newline
      \textit{Reward}: $1 - 0.9 * (step_{count} / max_{steps})$ for success or 0 for failure (Sparse). \\
    \midrule
    \includegraphics[width=1.5cm]{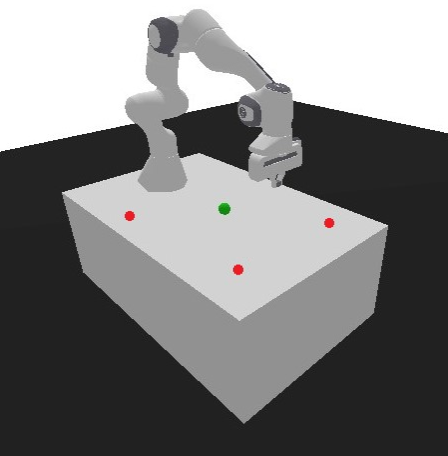} 
    & \raggedright\arraybackslash \textbf{Panda-Gym} 
    & \raggedright\arraybackslash \textit{Environments}: PandaReach environment. \newline \textit{Motivation}: Realistic 3D robotic control scenario. \newline
      \textit{States}: Robotic arm positions, velocities (Continuous). \newline \textit{Actions}: Joint torques for robotic arm (Continuous). \newline
      \textit{Reward}: Gradual reward based on distance to goal (Dense). \\
    \bottomrule
\end{tabular}
\caption{Comparison of domains, visualizations, and their specific characteristics.}
\label{tab:domains}
\end{table*}

\subsection{Online One-Shot Inference}  

When a new, partially observed trajectory $\trajectory_{obs}=(\astate_0,\action_0,\dots,\astate_T)$ arrives, we must quickly determine which $\policy_g$ best explains it.  
Rather than invoking planning or rolling out policies in the environment, we compute for each goal a single similarity score $
S_g(\trajectory_{obs}) \;=\; \ell\bigl(\trajectory_{obs};\pi_g\bigr),$ and then select $
\hat g \;=\;\arg\max_{g\in\goals} S_g(\trajectory_{obs})\,.$

To support a lightweight, one-shot scoring method that can be applied uniformly across all IL algorithms used in our experiments, we implement a similarity metric using \textbf{Negative Average MSE}.
\begin{align*}
  \ell_{\mathrm{MSE}}(\trajectory_{obs};\pi_g)
   = -&\frac{1}{T}\sum_{t=0}^{T-1}\bigl\|\pi_g(\astate_t) - \action_t\bigr\|^2, \\
  &
  S_g^{\mathrm{MSE}}(\trajectory_{obs}) = \ell_{\mathrm{MSE}}(\trajectory_{obs};\pi_g).
\end{align*}
This metric extends naturally to GAIL and AIRL by applying it to their policy heads $\pi_g$. 
Using this score, GRAIL performs fast inference over an entire trajectory in a single batch pass without requiring planning or RL rollouts.

% \frm[inline]{Idea for future research, try to detect biases (and the agent itself), by contrasting optimal policies with the IL-ones, and reason about this difference. }

\subsection{Training versus Inference Regimes}

BC-GRAIL trains one goal-conditioned policy per goal purely from offline demonstration trajectories, without interacting with the environment.
GAIL-GRAIL and AIRL-GRAIL follow standard adversarial imitation learning: during training they alternate between updating a discriminator and a policy using environment rollouts, so environment interaction occurs at training time on the same order as other RL/IL methods.
Crucially, at inference time, all GRAIL variants perform GR without any planner calls or environment interaction: they only evaluate the learned policies on the observed partial trajectory and compute the scoring function.

\section{Experimental Design}
\label{sec:exp1}

\paragraph{Environments}
We evaluate all methods on two domains with distinct characteristics: a discrete navigation environment -  MiniGrid ~\citep{MinigridMiniworld23} and a continuous robotic manipulation environment - PandaReach  \citep{gallouedec2021pandagym}. A summary of the differences appears in Table~\ref{tab:domains}. In the discrete MiniGrid task, the agent operates in a $9\times9$ grid, starting at position $(1,4)$ facing right. A fixed obstacle is placed at $(7,4)$, and possibly six candidate goals. %are located at $(7,1), (7,7), (5,1), (5,7), (7,3), (7,5)$. %The agent can choose from three discrete actions: \emph{turn left}, \emph{turn right}, and \emph{move forward}. Rewards are sparse and only provided upon reaching a goal.
In the continuous PandaReach manipulation task, a simulation of a robotic arm is trained to reach 3D targets, with possibly four candidate goals. %Observations include the end-effector's position and the velocities. The reward is a dense, smooth function based on the Euclidean distance between the end-effector and the target position.

In both environments, we manually defined the initial conditions (e.g., start position and orientation) and selected specific sets of candidate goals to construct controlled settings for GR (See Apendix \ref{sec:hyperparam}). This manual configuration allows us to systematically generate expert behaviors under different assumptions: (i) \textit{optimal} behavior, where the agent takes the shortest valid path to the goal, (ii) \textit{systematically-biased} behavior, where the agent follows a structurally preferred route (e.g., favoring straight-line motion or a specific route like in the example in Figure \ref{fig:nav}), and (iii) \textit{suboptimal} behavior, where the agent’s trajectory deviates due to noise or heuristic-driven choices. These settings are crucial for evaluating how well GR methods generalize under realistic, non-idealized behavior. We elaborate further on the generation of these behavior regimes in the next Section.

To further support stable IL across all experiments and algorithms, we standardized the episode structure by using fixed-horizon rollouts instead of early termination upon goal reaching. Specifically, GAIL and AIRL require the input demonstrations to have a fixed-horizon.
All experiments are seeded using the random number generator provided by Gymnasium~\cite{brockman2016openai}.

\paragraph{Baselines}  
We compare GRAIL against two recent GR methods that frame inference as a matching problem against goal‐directed policies.  
First, GRAQL~\cite{amado2022goal} learns a separate Q-function $Q_g(s,a)$ for each goal via Q-learning; at test time it treats the observed trajectory $\trajectory_{obs}$ as inducing a ``pseudo-policy'' over the taken actions and scores each goal by the negative sum of Kullback–Leibler (KL)-divergences between this pseudo-policy and the softmax policy derived from $Q_g$.  
Second, DRACO~\cite{nageris2024goal} trains each goal-directed policy $\policy_g$ via PPO (Proximal Policy Optimization)~\cite{schulman2017proximal} on the true reward and then compares the empirical distribution of $(s,a)$ pairs in $\trajectory_{obs}$ to the occupancy measure of $\policy_g$ using the 1-Wasserstein distance.  
Both baselines require per-goal RL training. 
 All hyperparameters, training details, information on computational resources and efficiency can be found in Appendix \ref{sec:hyperparam}.%\osher{I didn't have time to create the relevant sections as were in GDGR in the appendix, but I assume we need to cite them here}.

% \noindent \textbf{Algorithms.}
% \osher{What do you think about this paragraph?} We implemented imitation learning algorithms using \textit{Imitation} library~\cite{gleave2022imitation}. For GRAQL we follow the implementation details for Q-Learning. For DRACO we follow the implementation details for PPO and SAC.

\begin{table*}[t]
\small
\centering
\begin{tabular}{llcccc}
    \toprule
    \textbf{Goals} & \textbf{Obs} & \textbf{GRAQL} & \textbf{BC-GRAIL} & \textbf{GAIL-GRAIL} & \textbf{AIRL-GRAIL} \\
    \midrule
    2 & 10\% & \textbf{0.8333 ± 0.3416} & 0.8 ± 0.3055 & 0.5674 ± 0.0735 & 0.5333 ± 0.0762 \\
    2 & 20\% & \textbf{0.9 ± 0.2108} & \textbf{0.9 ± 0.2108} & 0.7886 ± 0.0494 & 0.7683 ± 0.0682 \\
    2 & 30\% & 0.9333 ± 0.2 & \textbf{1.0 ± 0.0} & 0.8652 ± 0.0306 & 0.8418 ± 0.0563 \\
    4 & 10\% & 0.4583 ± 0.2401 & \textbf{0.5875 ± 0.2286} & 0.5107 ± 0.0835 & 0.4784 ± 0.0733 \\
    4 & 20\% & 0.6833 ± 0.2236 & \textbf{0.725 ± 0.2092} & 0.7502 ± 0.0517 & 0.7223 ± 0.0662 \\
    4 & 30\% & 0.8125 ± 0.2734 & \textbf{1.0 ± 0.0} & 0.8328 ± 0.0349 & 0.8012 ± 0.0589 \\
    6 & 10\% & 0.3067 ± 0.1374 & 0.4222 ± 0.1089 & \textbf{0.4726 ± 0.0768} & 0.4450 ± 0.0652 \\
    6 & 20\% & 0.5111 ± 0.1764 & 0.6333 ± 0.1528 & \textbf{0.7101 ± 0.0531} & 0.6833 ± 0.0617 \\
    6 & 30\% & 0.5722 ± 0.2636 & \textbf{0.8444 ± 0.1736} & 0.7934 ± 0.0372 & 0.7651 ± 0.0571 \\
    \bottomrule
\end{tabular}
\caption{$F_1$-score (mean $\pm$ std over 10 GR tasks) comparison for \textbf{Suboptimal} trajectories with 2, 4, 6 goals under 10\%, 20\%, and 30\% observability. Bold indicates best performing method in each condition.}
\label{tab:subopt_f1}
\end{table*}

\paragraph{Behavioral Conditions}  
To probe robustness, we collect two types of demonstration data.  Under \emph{optimal} conditions, MiniGrid trajectories are handcrafted to follow the shortest path to each goal, while PandaReach trajectories are produced by a converged SAC (Soft Actor-Critic)~\cite{haarnoja2018soft} agent.  Under \emph{suboptimal} conditions, we introduce systematic deviations: in MiniGrid we insert 180° turns (left and right) before each action without changing the overall path. In PandaReach, we use two types of action noise - Gaussian noise at levels of 0.1 and 0.3, and uniform noise at levels of 0.1 and 0.3 - and evaluated all experiments under low observability (2\% of the trajectory).  For MiniGrid, we generate 10 trajectories per goal (using 7 for IL training and 3 for testing its GR performance); for PandaReach we collect 200 per goal (150/50 train/test split).
We evaluate GR performance, framed as a classification task, using standard metrics: Accuracy, Recall, Precision, and $F_1$-score.

\section{Results}
\label{subsec:results}
We tested GRAIL under various conditions to evaluate its performance, given systematically biased behavior (Section \ref{sec:biased}), suboptimal behavior (Section \ref{subsec:results_suboptimal}), and optimal behavior (Section \ref{subsec:results_optimal}). We further evaluate GRAIL's efficiency in terms of training requirements (Section \ref{sec:complexity}) and provide an ablation test for GRAIL's inference metric (Section \ref{sec:ablation}).

\begin{figure}[t]
    \centering
    \begin{subfigure}[t]{0.49\linewidth}
        \centering
        \includegraphics[width=\linewidth]{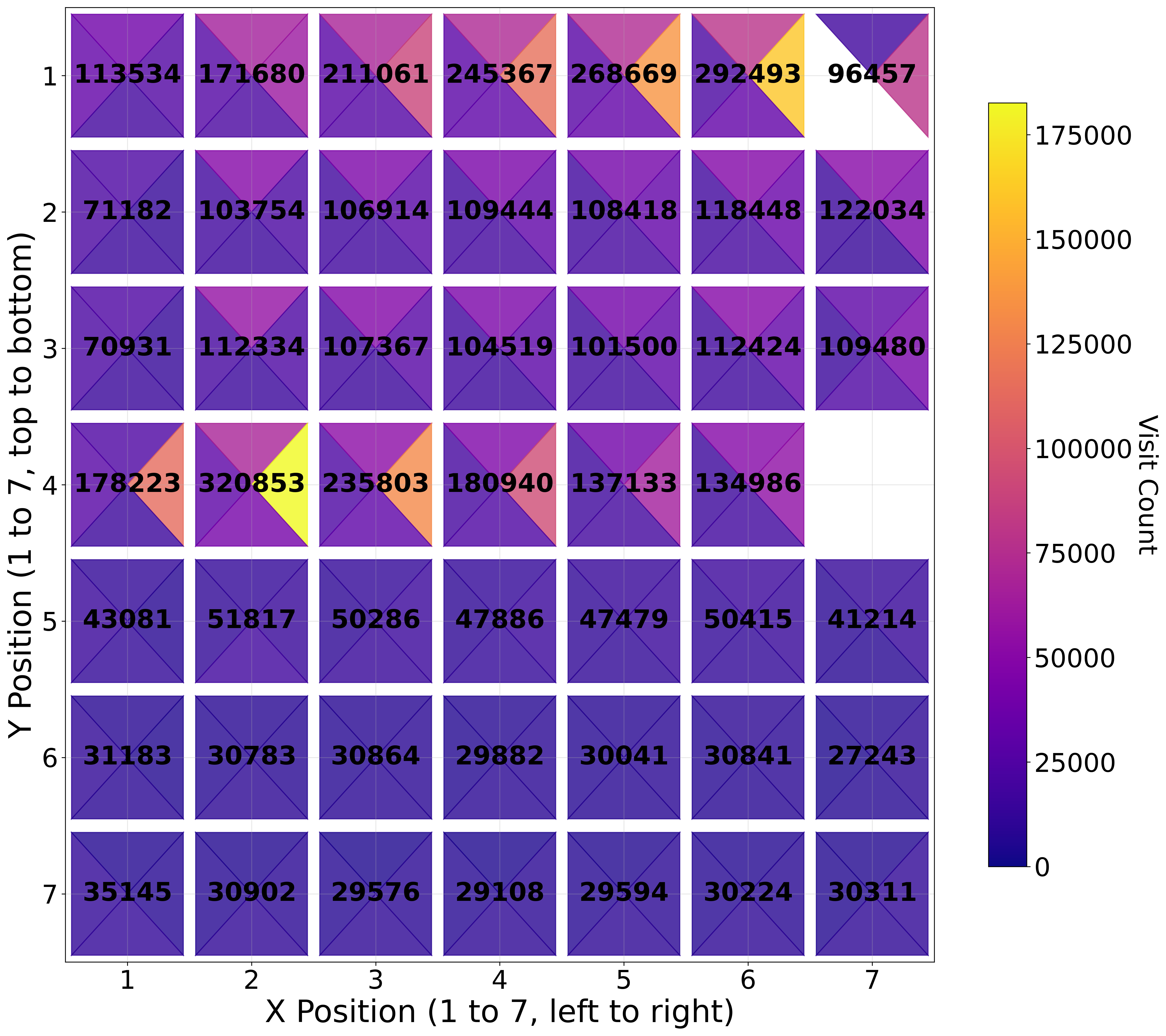}
        % \caption{Visit counts heatmap for Q-learning training toward goal (7,1).}
        \label{fig:heatmap_goal_7x1}
    \end{subfigure}
    \hfill
    \begin{subfigure}[t]{0.49\linewidth}
        \centering
        \includegraphics[width=\linewidth]{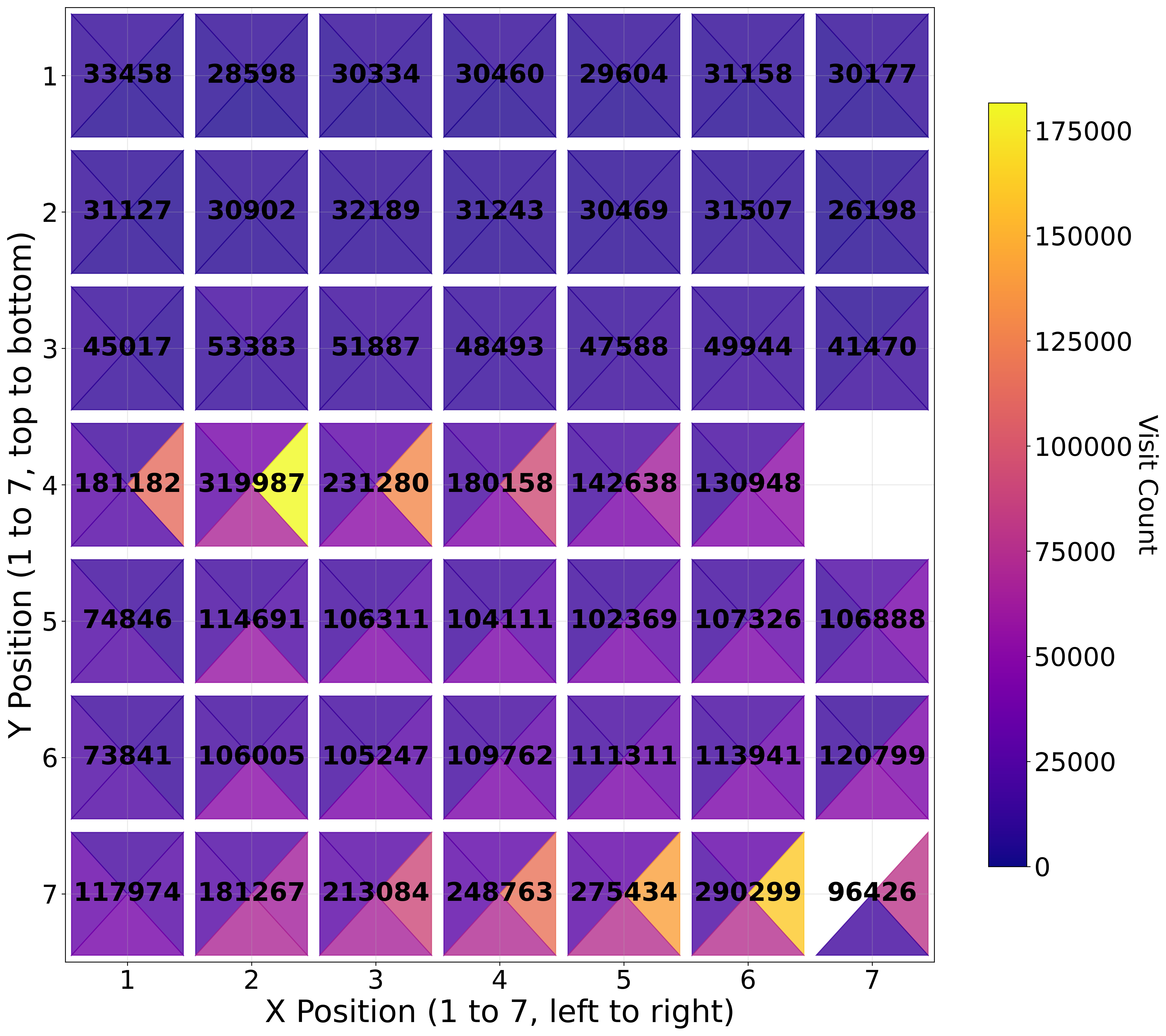}
        % \caption{Visit counts heatmap for Q-learning training toward goal (7,7).}
        \label{fig:heatmap_goal_7x7}
    \end{subfigure}
    \caption{
        Minigrid (9×9 grid, 7×7 free states) Q-learning visit counts per state (position + orientation) during training, shown as heatmaps with overlaid counts.  Yellow indicates high visit frequency.  (left) Training visits for goal at \((7,1)\).  (right) Training visits for goal at \((7,7)\).
    }
    \label{fig:minigrid_q_learning_heatmaps}
\end{figure}

Each reported result is averaged over 10 random seeds, with different initial environment states and policy initializations for each seed.
We report mean $\pm$ standard deviation for all metrics (accuracy, precision, recall, F$_1$-score).
For the main comparisons between GRAIL variants and baselines, we also compute 95\% confidence intervals for the mean (Appendix ~\ref{sec:detailed_stats}).
% In tables and figures, boldface marks the best performing method in each condition based on mean score.

\subsection{Systematically Biased Optimal Behavior}
\label{sec:biased}
In many scenarios, agents pursue goals along nominally optimal paths but exhibit consistent preferences for certain trajectories over equally short alternatives. We term this \emph{systematically biased optimal behavior}, where standard GR methods that assume uniform optimality can fail to distinguish goals when the biased preference obscures the distinguishing segments of the trajectory. This is the situation in the use-case in Figure ~\ref{fig:nav}, which we translated to an experimental setup in Minigrid: the two goals, \((7,1)\) and \((7,7)\), are placed in a 9×9 obstacle world. Each goal admits multiple shortest-path solutions, yet Q-learning training visit-counts (Fig.~\ref{fig:minigrid_q_learning_heatmaps}) hint that the optimal routes (``east then north'' for \((7,1)\), ``east then south'' for \((7,7)\)) dominate. We synthetically generated biased demonstrations in which the agent chooses its preferred optimal variant to be ``north then east'' for \((7,1)\), and ``east then south'' path for \((7,7)\).  

We collect ten biased trajectories per goal and compared GRAQL against the GRAIL variants (BC-GRAIL, GAIL-GRAIL, AIRL-GRAIL) learned goal‐directed policies.  Each method was evaluated on tasks in which only the first 20-40\% of each trajectory was observed, with performance quantified by Accuracy, Precision, Recall, and F$_1$‐score. 
%and applied two families of recognition methods.  First, the GRAQL baseline trained Q‐learning policies for each goal and performed KL‐divergence–based inference.  Second, the GRAIL variants (BC-GRAIL, GAIL-GRAIL, AIRL-GRAIL) learned goal‐directed policies.  Each method was evaluated on GR tasks in which only the first 20-40\% of each biased trajectory was observed, with performance quantified by Accuracy, Precision, Recall, and F$_1$‐score. 

\begin{figure}[t]
    \centering
    \includegraphics[width=1\linewidth]{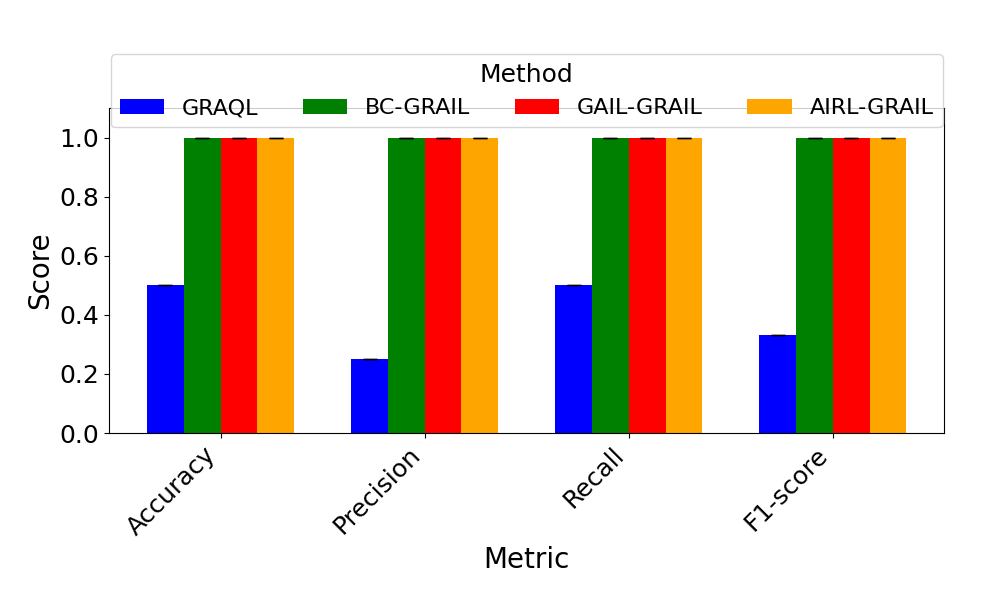}
    \caption{F$_1$-score comparison on Minigrid with systematically biased optimal demonstrations and 2 goals under 20–40\% observability across 10 runs. GRAIL variants (green/red/yellow) achieve perfect inference, compared to GRAQL (blue).}
    \label{fig:graql_grail_biased_optimal_comparison}
\end{figure}

As shown in Fig.~\ref{fig:graql_grail_biased_optimal_comparison}, all three GRAIL methods recover the true goal with 100\% F$_1$ across the entire 20–40\% observation range. In contrast, GRAQL's F$_1$ hovers around 0.3–0.5, effectively at chance, because the initial biased path segments (for (7x7) path) coincide for both goals and the Q‐functions assign them equal value. These results confirm that GRAIL, by directly modeling the biased distribution, can disambiguate goals even when RL‐based inference cannot.

% Next, we evaluate GRAQL and GRAIL using the three IL methods under both optimal and suboptimal behavior regimes. For each domain, we assess GR accuracy using F$_1$-score as a function of partial observability (i.e., the proportion of the demonstration revealed at test time) and the number of candidate goals. 
%

\begin{table*}[t]
\centering
\begin{tabular}{llcccc}
    \toprule
    \textbf{Goals} & \textbf{Obs} & \textbf{GRAQL} & \textbf{BC-GRAIL} & \textbf{GAIL-GRAIL} & \textbf{AIRL-GRAIL} \\
    \midrule
    2 & 10\% & \textbf{0.8 ± 0.3055} & 0.7333 ± 0.3266 & 0.7055 ± 0.0863 & 0.6604 ± 0.1092 \\
    2 & 20\% & \textbf{0.9333 ± 0.1491} & 0.8667 ± 0.2309 & 0.9194 ± 0.0407 & 0.9047 ± 0.0793 \\
    2 & 30\% & \textbf{1.0 ± 0.0} & 0.9333 ± 0.2 & 0.9649 ± 0.0230 & 0.9447 ± 0.0573 \\
    4 & 10\% & \textbf{0.7167 ± 0.2825} & 0.6333 ± 0.2485 & 0.6412 ± 0.0973 & 0.6032 ± 0.0851 \\
    4 & 20\% & \textbf{0.9 ± 0.1833} & 0.8167 ± 0.2191 & 0.8903 ± 0.0427 & 0.8453 ± 0.0701 \\
    4 & 30\% & \textbf{0.9667 ± 0.1} & 0.9333 ± 0.1333 & 0.9482 ± 0.0261 & 0.9331 ± 0.0610 \\
    6 & 10\% & 0.5078 ± 0.1584 & 0.3733 ± 0.1839 & \textbf{0.5772 ± 0.0934} & 0.5330 ± 0.0805 \\
    6 & 20\% & 0.8111 ± 0.2033 & 0.7111 ± 0.2129 & \textbf{0.8462 ± 0.0496} & 0.8147 ± 0.0693 \\
    6 & 30\% & 0.8694 ± 0.1417 & 0.7889 ± 0.1262 & \textbf{0.9188 ± 0.0291} & 0.8904 ± 0.0654 \\
    \bottomrule
\end{tabular}
\caption{$F_1$-score (mean $\pm$ std over 10 GR tasks) comparison for \textbf{Optimal} trajectories with 2, 4, 6 goals under 10\%, 20\%, and 30\% observability. Bold indicates best performing method in each condition.}
\label{tab:optimal_f1}
\end{table*}

\subsection{Suboptimal Behavior}
\label{subsec:results_suboptimal}

In real‐world settings, agents often deviate systematically from shortest‐path optima, whether through exploration, sensor noise, or habitual biases. We term this regime \emph{suboptimal behavior}. We begin this evaluation with MiniGrid by varying task complexity. We adjust the number of candidate goals and the visibility of the agent’s trajectory to evaluates how each method generalizes as the problem becomes harder. In this domain, we evaluate GR with 2, 4, and 6 candidate goals across three partial observability levels (10\%, 20\%, 30\% of the trajectory visible). Table~\ref{tab:subopt_f1} reports F$_1$-scores (mean ± std) for the Q-learning-based GRAQL baseline and for BC-, GAIL-, and AIRL-GRAIL. GRAQL’s performance degrades as the number of goals increases: with 6 goals at 10\% visibility, it achieves only 0.31±0.14, while BC-GRAIL reaches 0.42±0.11, GAIL-GRAIL 0.47±0.08, and AIRL-GRAIL 0.45±0.07. %Even with smaller goal sets, AIRL-GRAIL methods maintain F$_1$ above 0.75 at 20–30\% visibility, whereas GRAQL remains below 0.70.
Remarkably, BC-GRAIL holds a lead in most of the scenarios, and achieves perfect recognition (1.00±0.00) for 4 goals at 30\% observability, highlighting the effectiveness of supervised cloning under sub-optimality.

% \usepackage{subcaption} % Add this to your preamble if not already present

% \begin{figure}[t]
%     \centering
%     \begin{subfigure}[t]{0.48\linewidth}
%         \centering
%         \includegraphics[width=\linewidth]{images/inference_times.png}
%         \caption{Optimal Demonstrations.}
%         \label{fig:draco_grail_optimal_comparison}
%     \end{subfigure}
%     \hfill
%     \begin{subfigure}[t]{0.48\linewidth}
%         \centering
%         \includegraphics[width=\linewidth]{images/panda_gr.png}
%         \caption{Suboptimal Demonstrations.}
%         \label{fig:draco_grail_suboptimal_comparison}
%     \end{subfigure}
%     \caption{$F_1$–score (± std across 10 GR tasks) comparison on PandaReach between BC-GRAIL and DRACO with optimal (left) and suboptimal (right) demonstrations under varying observability levels.}
%     \label{fig:combined_panda_results}
% \end{figure}

\begin{figure}[t]
    \centering
    \includegraphics[width=1\linewidth]{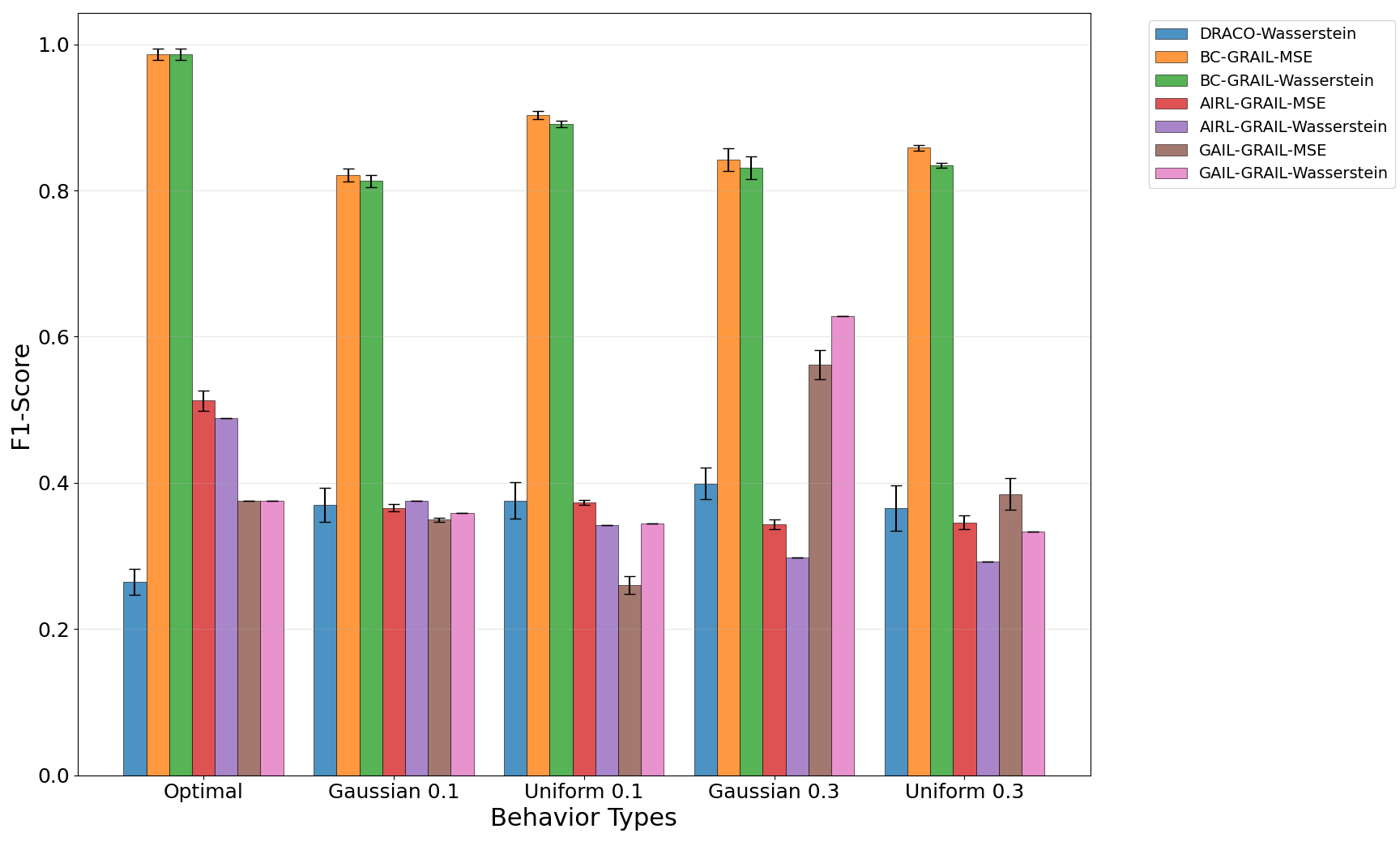}
    \caption{$F_1$-score comparison on PandaReach with optimal and noisy demonstrations (uniform and gaussian) and 4 goals under 2\% observability across 10 runs.}
    \label{fig:draco_vs_grail_gr_panda}
\end{figure}

In PandaReach, we compare GRAIL methods against the DRACO baseline in a four‐goal continuous‐control task under 2\% observability and with Gaussian and uniform action noise at 10\% and 30\%. Figure~\ref{fig:draco_vs_grail_gr_panda} illustrates that BC‐GRAIL outperforms all other methods across noise levels, while GAIL‐GRAIL shows particular strength as noise increases. AIRL‐GRAIL and DRACO lag behind in these noisy settings. These results confirm that learning from suboptimal demonstrations enables BC‐based policies to internalize systematic deviations, whereas approaches predicated on near-optimal behavior treat such deviations as noise.

% In PandaReach, we compare BC‐GRAIL against the DRACO baseline in a 4‐goal continuous‐control task under the same 10\% random‐action corruption. Figure~\ref{fig:draco_grail_suboptimal_comparison} illustrates the results at 10\% and 50\% observability. At 10\%, DRACO’s F$_1$ collapses to 0.09±0.004, reflecting its brittleness to sparse and noisy data; BC‐GRAIL, by contrast, achieves 0.58±0.03 by internalizing the precise noise distribution. At 50\%, DRACO recovers to 0.65±0.002 (much better than random), yet BC‐GRAIL outperforms it at 0.78±0.05. These results confirm that learning directly from suboptimal demonstrations allows GRAIL policies to capture systematic deviations, whereas methods assuming near‐optimality treat them as uninformative noise.

% This robustness arises because GRAIL variants learn directly from the noisy or biased demonstrations, rather than treating deviations as random noise, thereby capturing systematic suboptimal patterns that Q‐learning–based inference cannot.

Together, these results demonstrate that GRAQL and DRACO are advantageous only when the demonstrations are perfect, or when the goal set is very small. In all other scenarios: moderate observability, larger goal sets, or suboptimal/expert‐imperfect data, GRAIL outperforms these methods. BC‐GRAIL’s surprisingly strong performance under suboptimal trajectories further highlights the value of BC when behavior noise is structured.
%These experiments highlight a critical weakness of Q‐learning–based inference: its reliance on optimality renders it incapable of modeling real‐world biases. In contrast, GRAIL’s imitation‐learning backends adapt seamlessly to structured noise, with adversarial IL delivering particularly robust performance and BC‐GRAIL demonstrating remarkable resilience under moderate complexity.

\subsection{Optimal Behavior}
\label{subsec:results_optimal}

When demonstrations are noise‐free shortest‐path trajectories, we refer to this as the \emph{optimal behavior} regime. Here, inference methods that assume idealized rationality should excel, given sufficient observation.

In the MiniGrid domain, Table~\ref{tab:optimal_f1} highlights several key patterns in F$_1$‐score performance under optimal demonstrations. When only 10\% of each trajectory is observed, GRAQL retains a modest advantage for small goal sets (2 and 4 goals), reflecting its reliance on precisely learned Q‐values for very sparse data. As we increase observability to 20\%, the gap narrows: GRAIL’s imitation‐based methods rapidly catch up. Notably, for the six‐goal scenario %—even at 10\% visibility—GAIL‐
GRAIL (0.58±0.09) and AIRL‐GRAIL (0.53±0.08) 
surpass GRAQL (0.51±0.16), indicating that adversarial IL’s enforced coverage of diverse expert behaviors yields policies whose early‐trajectory predictions generalize better across larger goal sets. By 30\%, GAIL‐GRAIL achieves 0.92±0.03 and AIRL‐GRAIL 0.89±0.07 for six goals, whereas GRAQL lags at 0.87±0.14. These results underscore that, while Q‐learning–based inference excels in simple tasks, imitation‐learning backends, especially those trained adversarially, scale more gracefully in data availability and goal‐complexity.  

% In MiniGrid, Table~\ref{tab:optimal_f1} presents F$_1$‐scores for 2, 4, and 6 goals under 10\%, 20\%, and 30\% partial observability. As expected, GRAQL holds a small lead in the most data‐scarce scenarios. However, by 20\% visibility all imitation‐based methods converge rapidly. At 30\% observability, BC‐GRAIL, GAIL‐GRAIL and AIRL‐GRAIL achieve 0.93±0.20, 0.96±0.02, and 0.94±0.06 respectively, equaling GRAQL’s perfect 1.00. For larger goal sets (4 and 6 goals), AIRL variants even overtake GRAQL at the lowest visibility.

In PandaReach, all GRAIL variants surpassed DRACO when using SAC‐generated optimal trajectories. At just 2\% observability (Figure~\ref{fig:draco_vs_grail_gr_panda}), DRACO’s performance falls well below that of each GRAIL method. This demonstrates that even under ideal conditions, imitation learning can match or outperform actor‐critic–based goal recognition when only limited data is available.

Results in this optimal demonstration regime confirms that while RL-based GR methods can yield strong performance under extreme data scarcity and small goal sets, GRAIL’s policies scale more gracefully as observation coverage and problem complexity grow. %AIRL’s built‐in state‐action coverage and BC’s efficiency make them compelling alternatives even when the assumptions of perfect rationality hold.

\begin{figure}[t]
    \centering
    \includegraphics[width=1\linewidth]{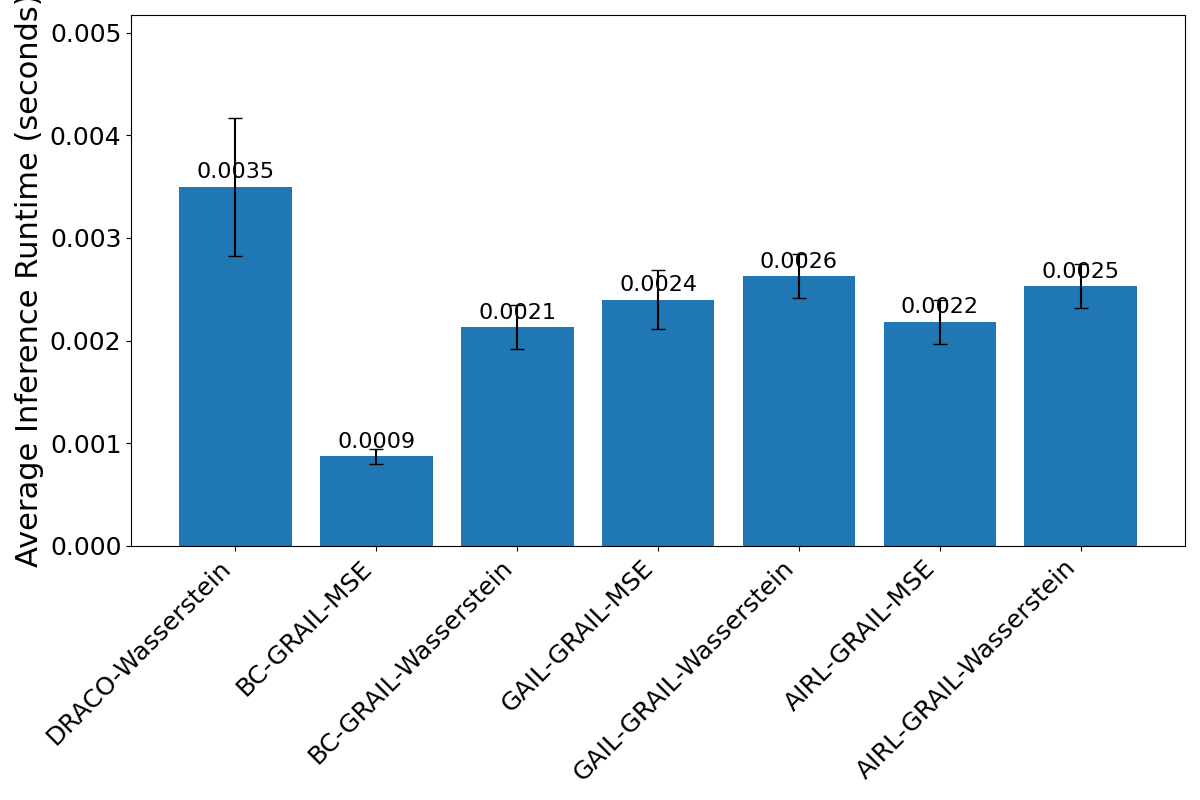}
    \caption{GR inference time comparison on PandaReach with optimal and noisy demonstrations (uniform and gaussian) and 4 goals across 10 runs.}
    \label{fig:draco_vs_grail_inference_time}
\end{figure}

\subsection{Efficiency and Simulation Complexity}  
\label{sec:complexity}

Beyond recognition accuracy, GRAIL delivers substantial efficiency gains both in training and inference.  In the MiniGrid domain, BC-GRAIL requires only about 5 s per goal, GAIL-GRAIL about 60 s, and AIRL-GRAIL about 240 s, compared to roughly 300 s for the GRAQL baseline.  In PandaReach, BC, GAIL, and AIRL each train in under half the time needed by the PPO-based DRACO recognizer.  Furthermore, BC-GRAIL achieves near-optimal recognition with just 7 demonstrations per goal in MiniGrid (and 150 in PandaReach), whereas GRAQL must visit on the order of $2\times10^5$ state–action pairs to converge its Q–functions (see the heatmap in Fig.~\ref{fig:minigrid_q_learning_heatmaps}).  At inference time, all GRAIL methods require only a handful of forward passes and are dramatically faster than DRACO (Fig.~\ref{fig:draco_vs_grail_inference_time}).

\textbf{Scalability to larger goal sets.}
The detailed experiments reported above use relatively small closed sets of candidate goals (up to 6 in MiniGrid and up to 4 in PandaReach), consistent with standard GR formulations that assume a finite set of candidate goals or hypotheses~\cite{meneguzzi2021survey,amado2022goal, nageris2024goal}.
%We also conducted additional internal tests with larger goal sets (higher $|G|$ in PandaGym, e.g. $|G|=8$) and observed qualitatively similar trends: absolute performance decreases for all methods as $|G|$ grows, but the relative advantage of GRAIL over RL-based baselines remained stable. 
We leave a more systematic large or infinite-$|G|$ study for future work.

\subsection{Inference Metric Ablation}
\label{sec:ablation}
We adopt negative average MSE between the policy's predicted actions and the observed actions as our default goal score because it is simple, numerically stable, and applies well to both discrete and continuous action spaces.
To situate this choice, we compare it against the Wasserstein-based distance proposed in DRACO~\citep{nageris2024goal}, which was shown to be a strong metric for policy-based GR in both MiniGrid and Panda-Gym, especially under partial observability and noisy observations, and to perform competitively with their Z-score distance.
Earlier GR-as-RL work (GRAQL)~\citep{amado2022goal} introduced three discrete, Q-based distances--MaxUtil, KL-divergence, and Divergence Point--and found that KL and MaxUtil generally dominate DP across a wide range of observability levels. Therefore we compare GRAIL to GRAQL with KL-divergence

In contrast to value-based distances such as MaxUtil, KL, Divergence Point, and $Z$-score, both our per-step MSE and DRACO's Wasserstein distance operate directly on the action distributions of goal-directed policies and apply uniformly in discrete and continuous action spaces. We therefore focus our comparison on Wasserstein as a representative policy-based metric: DRACO empirically compares the GRAQL-style distances (MaxUtil, KL, and Divergence Point) against Wasserstein and $Z$-score, and reports that the Wasserstein- and $Z$-score–based variants achieve superior GR accuracy in most evaluated settings~\citep{nageris2024goal}. %Among metrics that share this policy-based, domain-agnostic nature, MSE and Wasserstein are thus the most natural pair to study in our ablations.}

% In contrast to these measures, both our per-step MSE and DRACO's Wasserstein distance operate directly on the action distributions of goal-directed policies and extend naturally to continuous action spaces.

% Concretely, our ablations (Section~\ref{sec:ablation}) compare MSE to a Wasserstein score as in DRACO: both metrics yield very similar GR accuracy across MiniGrid and PandaReach, but MSE is cheaper to compute at inference time and slightly more numerically stable when policies allocate very low probability mass to observed actions.
% For this reason, we use MSE as the default scoring metric in GRAIL, and treat Wasserstein as a principled, policy-based baseline grounded in prior GR work.
% GRAIL handles suboptimal or systematically biased behaviors implicitly, as the learned policies are trained directly on such behavior.

We measure both inference latency and recognition accuracy across all GRAIL variants using both our proposed negative averaged‐MSE scoring function, and Wasserstein distance.
% \textbf{MiniGrid (discrete actions).}
% For the discrete-action MiniGrid domain, we compare MSE against a cross-entropy (log-likelihood) score over the policy's categorical distribution: $S_g^{\text{CE}}(\trajectory_{obs}) = \frac{1}{T}\sum_{t=0}^{T-1}\log\pi_g(\action_t\mid\astate_t)$.
% This is the natural probabilistic score for discrete action policies.
% Empirically, we find that MSE and cross-entropy yield similar goal recognition performance across optimal, suboptimal, and biased demonstration regimes, with MSE being slightly more stable under high noise.
% Since MSE avoids numerical issues with log-probabilities (e.g., when $\pi_g(\action_t\mid\astate_t)$ is near zero) and is cheaper to compute, we adopt it as the default metric.
% \textbf{PandaReach (continuous actions).}
% For continuous-action tasks, we compare MSE against a Gaussian log-likelihood score, which models $\pi_g(\cdot\mid\astate_t)$ as a Gaussian with mean $\mu_g(\astate_t)$ and fixed covariance $\Sigma$: $S_g^{\text{Gauss}}(\trajectory_{obs}) = \frac{1}{T}\sum_{t=0}^{T-1}\log\mathcal{N}(\action_t\mid\mu_g(\astate_t),\Sigma)$.
Figure~\ref{fig:draco_vs_grail_inference_time} plots the per‐trajectory scoring time for each method under both MSE and Wasserstein metrics.  Across the board, MSE‐based scoring reduces latency, and this speed‐up accumulates when evaluating large goal sets.
Figure~\ref{fig:draco_vs_grail_gr_panda} compares F$_1$‐scores under both similarity metrics in PandaReach, across four observation noise regimes (optimal, Gaussian noise at 0.1 and 0.3, and uniform noise at 0.1 and 0.3). Under perfect demonstrations, BC-GRAIL performs equally well with either metric, while AIRL-GRAIL sees a modest gain when using MSE‐based scoring.  As noise increases, BC-GRAIL consistently benefits from MSE, showing improved robustness, and AIRL-GRAIL likewise enjoys moderate gains.  By comparison, GAIL-GRAIL retains a clear advantage under the Wasserstein metric across noisy settings.

\section{Related Work}

\textbf{Planning-based GR.}
The foundational ~\citet{ramirez2009plan} framework casts GR as computing goal likelihoods via domain-independent planning, assuming known domain models.
Subsequent work extended this to landmark-based inference~\cite{pereira2020landmark}, with a comprehensive survey by~\citet{meneguzzi2021survey}. This work was extended by GRNet to leverage a learning paradigm over the PDDL representation \cite{chiari2023goal}.

\textbf{RL-based GR.}
Goal Recognition as Q-Learning (GRAQL)~\cite{amado2022goal} learns per-goal Q-functions via Q-learning and scores observations by comparing them to the softmax policies derived from these value functions, removing the need for explicit domain models.
More recent work~\cite{nageris2024goal, fang2023real} trains deep policy networks using actor-critic methods and performs recognition via Wasserstein distance, extending to continuous control domains. Both in planning-based approaches and RL-based ones, there is an assumption that the actor's behavior in the environment is consistent with the planner/learner's policy.

% \textcolor{blue}{\textbf{Online and dynamic GR.}
% Online GR methods handle streaming observations, enabling incremental inference as new data arrives~\cite{vered2017heuristic}.
% Recent work includes Online Dynamic Goal Recognition (ODGR)~\cite{shamir2024odgr}, which adapts to evolving goal sets and environment dynamics.}

GRAIL differs from these approaches by learning per-goal policies via IL rather than via planning or pure RL on environment rewards, thus it can better capture biases and suboptimal patterns that emerge in the actor's behavior we aim to recognize.
It performs planner-free, one-shot scoring of partial trajectories using these learned policies, designed to account for suboptimal and systematically biased agent behavior. %while remaining competitive under optimal behavior.
GRAIL handles both discrete and continuous environments, demonstrating that imitation-based policy learning provides a flexible, data-driven alternative to planning- and RL-based GR methods.

%\correct{We now continue to discuss specific IL methods which we leverage in this work for learning goal-directed policies.}{}{Faff. Ohterwise, you may wish to write a connecting bit in Section~\ref{sec:method}}

\section{Conclusion}
\label{sec:conclusion}

This paper introduces GRAIL (Goal Recognition Alignment through Imitation Learning), a framework for inferring an agent's goal from partial observations of its behavior given a finite set of candidate goals. The key idea is to learn one goal-directed policy per candidate goal using imitation learning methods such as Behavioral Cloning (BC), Generative Adversarial Imitation Learning (GAIL), and Adversarial Inverse Reinforcement Learning (AIRL), instead of relying on planners or environment reward functions. At test time, GRAIL performs planner-free, one-shot goal recognition (GR) by scoring how well each goal-directed policy explains the observed partial trajectory, without any environment rollouts. This yields a unified way to apply imitation learning (IL), including AIRL as a special case, to GR tasks in both discrete (MiniGrid) and continuous (PandaReach) control domains.

We evaluate GRAIL in settings with optimal, noisy, and systematically biased behavior. Across MiniGrid and PandaReach benchmarks, GRAIL variants often match or outperform RL-based GR baselines such as GRAQL and DRACO in recognition accuracy and F$_1$-score, particularly at low observability. An ablation on scoring metrics compares negative mean squared error (MSE) against Wasserstein score and shows that MSE achieves a favorable trade-off between accuracy and inference time while remaining simple and numerically stable. %We also report mean and standard deviation over multiple seeds, alongside confidence intervals, to support the reported performance differences.

Beyond GR accuracy and run-time, GRAIL offers several system‐level attributes that are critical for practitioners. It provides \emph{flexibility}: different IL backends can be swapped in depending on data availability and domain characteristics, for example BC in supervision-rich domains or GAIL and AIRL when broader generalization is needed. Second, GRAIL is inherently \emph{modular}: after the offline training of per-goal policies, online inference requires only a single batch of forward pass, with no additional environment interaction or supervision.
Moreover, GRAIL achieves \emph{robustness} by aligning directly with observed behaviors, rather than penalizing every deviation from an idealized optimal planner/policy. Finally, because inference uses only learned policies and a simple scoring rule, the framework is easy to integrate into larger multiagent systems that need fast GR aligned with observed behavior.

\subsection*{Limitations and Future Work}
Despite these strengths, GRAIL has several limitations that point to directions for future work. First, the framework assumes a closed, known set of candidate goals. This is the standard formulation in much of the GR literature, where the task is to classify observed behavior among a finite set of hypotheses. Open-set GR, where goals can lie outside a predefined inventory or must be discovered online, requires additional tools such as novelty detection, goal discovery, or dynamic hypothesis generation \cite{shamir2024odgr}. Extending GRAIL to handle unknown or changing goal sets is an important next step.

Second, our experiments assume fixed environment dynamics. We do not study severe shifts in dynamics, changes in action semantics, or transfer to entirely new environments. Understanding how to adapt or regularize goal-conditioned policies so that GRAIL remains effective under such shifts is an important future direction.

Third, the empirical study focuses on simulated domains with vector observations and synthetic models of suboptimality and bias. These settings make it possible to control noise and structure, but they do not capture complex, raw inputs from for perception-heavy tasks, which are usually attributed to the complementary problem of \textit{Activity Recognition} rather than \textit{Goal Recognition} \cite{sukthankar2011activity, mirsky2021introduction}.

%Applying GRAIL to richer observation modalities, such as images, videos, or sensor streams, would require combining the current framework with representation learning and addressing issues of sample efficiency and generalization.}

Fourth, per-goal training scales linearly with the number of candidate goals. Although this is acceptable for the inventories considered here and is common in many GR approaches, very large goal sets would increase training cost and storage demands. Future work could investigate ways to learn general goal-conditioned policies that support zero-shot or few-shot transfer learning for new goals without retraining a policy from scratch~\cite{elhadad2025general, shamir2024odgr}.

Finally, we explore only a limited set of policy architectures and scoring functions. More expressive policies, attention mechanisms, or sequence models might capture structured biases more effectively, and calibrated probabilistic scores could provide better uncertainty estimates over goals. Investigating these extensions, while retaining the simple and planner-free inference pipeline, is a natural progression of this line of work.

\subsection*{Societal and Ethical Implications for AI Alignment} 
Interpreting an agent's goal from behavior is a core challenge in AI alignment, but doing so without imposing unrealistic assumptions about optimality or rationality raises significant ethical and policy considerations. While GRAIL enables more behaviorally-grounded goal inference, particularly for humans whose actions reflect preferences and constraints beyond normative models, this capability comes with risks. Systems that infer goals from demonstrations must be designed with strong safeguards to prevent misuse, such as privacy violations, surveillance, or manipulation of inferred intent \cite{kautz2022third}. As GR becomes increasingly integrated into human-facing applications, it is essential to uphold the following invariants: (1) transparency in how inferences are generated and communicated to users \cite{chakraborti2017plan, wagner2021explanation, winfield2021ieee}, (2) privacy protection by minimizing the collection and retention of sensitive behavioral data \cite{cai2024fedhip}, (3) fairness by preventing systematic biases in GR across different user groups, and (4) user agency by enabling individuals to understand, contest, or override inferred goals \cite{coman2018ai, kasenberg2018norm, mirsky2025artificial}. While GRAIL offers novel approach for aligned GR, its deployment must be governed by policies that preserve user values and protect against exploitative use.

\clearpage

\bibliographystyle{plainnat}
\bibliography{aaai2026}

\clearpage 
% \input{reproducability}

% \clearpage

\appendix

\clearpage

\section{Experimental Setup}
\label{sec:hyperparam}

\subsection{Domains}
Our experiments span two domains: \textbf{MiniGrid} (a discrete gridworld) and \textbf{PandaReach} (a continuous-control robotic arm environment). Each domain enables evaluation under both optimal and non-optimal agent behaviors.

\subsection{Algorithms and Hyperparameters}

All three IL algorithms - BC, GAIL, and AIRL - were implemented using the \texttt{Imitation} library\footnote{\url{https://imitation.readthedocs.io/}}, following its standard structure and practices. Below, we detail the primary hyperparameter configurations, along with alternative values explored. Multiple reasonable configurations were tested (over several random seeds), but the selected values yielded the best and most consistent performance.

\paragraph{Behavioral Cloning (BC):}
\begin{itemize}
    \item \texttt{batch\_size}: $8$ \hspace{2em} [other values tried: $16$, $32$]
    \item \texttt{learning\_rate}: $0.001$ \hspace{2em} [other values tried: $0.0005$, $0.005$]
\end{itemize}

\paragraph{GAIL:}
\begin{itemize}
    \item \texttt{gen\_alg}: PPO
    \item \texttt{policy}: "MlpPolicy"
    \item \texttt{batch\_size}: $32$ \hspace{2em} [other values tried: $64$, $128$]
    \item \texttt{gamma}: $0.95$ \hspace{2em} [other values tried: $0.99$, $0.90$]
    \item \texttt{learning\_rate}: $0.0003$ \hspace{2em} [other values tried: $0.0005$, $0.0001$]
    \item \texttt{n\_disc\_updates\_per\_round}: $8$
    \item \texttt{gen\_replay\_buffer\_capacity}: $512$
    \item \texttt{demo\_batch\_size}: $16$
\end{itemize}

\paragraph{AIRL:}
\begin{itemize}
    \item \texttt{gen\_alg}: PPO
    \item \texttt{policy}: "MlpPolicy"
    \item \texttt{batch\_size}: $32$ \hspace{2em} [other values tried: $64$, $128$]
    \item \texttt{gamma}: $0.95$ \hspace{2em} [other values tried: $0.99$, $0.90$]
    \item \texttt{learning\_rate}: $0.0003$ \hspace{2em} [other values tried: $0.0005$, $0.0001$]
    \item \texttt{n\_disc\_updates\_per\_round}: $8$
    \item \texttt{gen\_replay\_buffer\_capacity}: $512$
    \item \texttt{demo\_batch\_size}: $16$
\end{itemize}

\subsection{MiniGrid Experiments}

\paragraph{Biased-Optimal Demonstrations}
The agent operates in a $9\times9$ MiniGrid environment, starting at position $(1,4)$ facing right. A fixed obstacle is placed at $(7,4)$. For each candidate goal, we generate 10 demonstration trajectories (using 7 for IL training and 3 for evaluating GR performance). Biased demonstrations are synthetically generated so that for the goal at $(7,1)$, the agent prefers the ``north-then-east'' path, while for $(7,7)$, it prefers ``east-then-south''. We compare GRAQL with the GRAIL variants (BC-GRAIL, GAIL-GRAIL, AIRL-GRAIL) using learned goal-directed policies. Performance is evaluated at varying observability levels (20\%--40\% of each trajectory visible) using Accuracy, Precision, Recall, and F$_1$-score.

\paragraph{Suboptimal Demonstrations}
In the suboptimal setting, we introduce stochastic 180$^\circ$ turns (left and right) before each action, which do not change the overall optimal path but introduce noise. We adjust both the number of candidate goals and trajectory observability to evaluate generalization under increasing difficulty. Specifically, GR is evaluated with 2, 4, and 6 candidate goals: 
\begin{itemize}
    \item 2-goal: $[(7,1), (7,7)]$
    \item 4-goal: $[(7,1), (7,7), (7,3), (7,5)]$
    \item 6-goal: $[(7,1), (7,7), (5,1), (5,7), (7,3), (7,5)]$
\end{itemize}
We also vary the observed trajectory fraction: 10\%, 20\%, 30\%. For each, we collect 10 demonstration trajectories per goal (7 train / 3 test).

\paragraph{Optimal Demonstrations}
The optimal demonstration setup in MiniGrid uses the same protocol as above but only with strictly optimal (shortest-path) agent behavior.

\subsection{PandaReach Experiments}

\paragraph{Suboptimal Demonstrations}
For PandaReach, we collect 200 trajectories per goal (150 for IL training, 50 for evaluation). Two types of action noise are used to generate suboptimal behaviors:
\begin{itemize}
    \item \textbf{Gaussian noise:} At each time step, Gaussian noise (with zero mean and standard deviation $\sigma=0.1$ or $\sigma=0.3$) is added to the agent's actions.
    \item \textbf{Uniform noise:} At each time step, uniform noise in the range $[-0.1, 0.1]$ or $[-0.3, 0.3]$ is added to the actions.
\end{itemize}
All experiments are conducted under low observability (2\% of each trajectory available for inference). In this domain, we compare GRAIL methods against the DRACO baseline in a four-goal continuous-control task, under each noise condition.

\paragraph{Optimal Demonstrations}
For PandaReach, the optimal demonstration protocol mirrors the suboptimal case, but with noise-free (optimal) trajectories.

\begin{table*}[ht]
\centering
\small
\begin{tabular}{lccc}
\toprule
\textbf{Package/Tool} & \textbf{Version} & \textbf{License} & \textbf{Citation/Link} \\
\midrule
minigrid      & 2.5.0  & Apache 2.0 & \citep{MinigridMiniworld23} \\
panda-gym     & 3.0.7  & MIT        & \citep{gallouedec2021pandagym} \\
DRACO code    & -      & Academic & \citep{nageris2024goal} \\
GRAQL code    & -      & Academic & \citep{amado2022goal} \\
Stable Baselines3 & 2.1.0 & MIT      & \url{https://stable-baselines3.readthedocs.io/} \\
Imitation       & 1.0.1  & MIT        & \url{https://imitation.readthedocs.io/} \\
\bottomrule
\end{tabular}
\caption{Summary of key external packages, versions, and licenses.}
\end{table*}

\subsection{Computational Resources and Efficiency}
\label{sec:compute}

This section summarizes the computational requirements and hardware used for all experiments. While training imitation and reinforcement learning models is resource-intensive, GR inference with trained policies is computationally light and can be executed in under one second on a standard CPU by simply loading the trained model.

\paragraph{Compute Settings.}
All training and evaluation procedures were performed on an Apple M1 Max machine with the following specifications:
\begin{itemize}
    \item \textbf{CPU:} Apple M1 Max, 10 cores (8 performance + 2 efficiency)
    \item \textbf{Memory:} 64 GB RAM
    \item \textbf{System Firmware Version:} 10151.140.19
    \item \textbf{OS Loader Version:} 10151.140.19
\end{itemize}
No discrete GPU was used; all experiments were conducted on CPU.

\paragraph{Software Stack.}
All code was run under \texttt{macOS 14.3 (Sonoma)}, using Python 3.12. Libraries and their versions are listed in Section~\ref{sec:licenses}.

\paragraph{Storage Requirements.}
The disk space required for saving trained models and supporting data varies by method:
\begin{itemize}
    \item \textbf{AIRL-GRAIL:} $\approx 800$ MB
    \item \textbf{GAIL-GRAIL:} $\approx 800$ MB
    \item \textbf{BC-GRAIL:} $\approx 400$ MB
    \item \textbf{DRACO:} $\approx 200$ MB
    \item \textbf{GRAQL:} $\approx 3000$ MB
\end{itemize}
GRAQL's high storage demand results from its use of tabular Q-learning, requiring a large Q-table. In contrast, DRACO and the GRAIL variants only store neural network weights, leading to significantly lower storage requirements.

\paragraph{Compute Efficiency.}
Once trained, all GR methods achieve inference times under one second per episode on CPU, with no need for further optimization or parallelization.

\subsection{Detailed Statistical Results}
\label{sec:detailed_stats}

This appendix provides complete statistical details for the main experimental results, including 95\% confidence intervals computed using the t-distribution (appropriate for $n=10$ seeds). All confidence intervals were computed from per-seed F$_1$-scores.

\subsubsection{MiniGrid Suboptimal Trajectories}

Table~\ref{tab:subopt_f1_detailed} presents mean $\pm$ standard deviation with 95\% confidence intervals for each experimental condition.

\begin{table*}[h]
\centering
\small
\begin{tabular}{llcccc}
    \toprule
    \textbf{Goals} & \textbf{Obs} & \textbf{GRAQL} & \textbf{BC-GRAIL} & \textbf{GAIL-GRAIL} & \textbf{AIRL-GRAIL} \\
    \midrule
    2 & 10\% & 0.83±0.34 & 0.80±0.31 & 0.57±0.07 & 0.53±0.08 \\
      &      & [0.55, 1.00] & [0.57, 0.93] & [0.51, 0.62] & [0.48, 0.59] \\
    2 & 20\% & 0.90±0.21 & 0.90±0.21 & 0.79±0.05 & 0.77±0.07 \\
      &      & [0.73, 0.99] & [0.73, 1.00] & [0.75, 0.83] & [0.72, 0.82] \\
    2 & 30\% & 0.93±0.20 & 1.00±0.00 & 0.87±0.03 & 0.84±0.06 \\
      &      & [0.76, 0.98] & [1.00, 1.00] & [0.84, 0.89] & [0.80, 0.88] \\
    4 & 10\% & 0.46±0.24 & 0.59±0.23 & 0.51±0.08 & 0.48±0.07 \\
      &      & [0.28, 0.64] & [0.42, 0.76] & [0.45, 0.57] & [0.42, 0.53] \\
    4 & 20\% & 0.68±0.22 & 0.73±0.21 & 0.75±0.05 & 0.72±0.07 \\
      &      & [0.52, 0.83] & [0.57, 0.88] & [0.71, 0.79] & [0.67, 0.77] \\
    4 & 30\% & 0.81±0.27 & 1.00±0.00 & 0.83±0.03 & 0.80±0.06 \\
      &      & [0.60, 0.97] & [1.00, 1.00] & [0.81, 0.86] & [0.76, 0.85] \\
    6 & 10\% & 0.31±0.14 & 0.42±0.11 & 0.47±0.08 & 0.45±0.07 \\
      &      & [0.20, 0.41] & [0.34, 0.50] & [0.41, 0.53] & [0.40, 0.49] \\
    6 & 20\% & 0.51±0.18 & 0.63±0.15 & 0.71±0.05 & 0.68±0.06 \\
      &      & [0.38, 0.64] & [0.52, 0.75] & [0.67, 0.75] & [0.64, 0.73] \\
    6 & 30\% & 0.57±0.26 & 0.84±0.17 & 0.79±0.04 & 0.77±0.06 \\
      &      & [0.38, 0.76] & [0.71, 0.96] & [0.77, 0.82] & [0.72, 0.81] \\
    \bottomrule
\end{tabular}
\caption{Detailed F$_1$-score statistics for suboptimal trajectories. Each cell shows mean±std in the first row and [95\% CI lower, upper] in the second row.}
\label{tab:subopt_f1_detailed}
\end{table*}

\subsubsection{MiniGrid Optimal Trajectories}

Table~\ref{tab:optimal_f1_detailed} presents mean $\pm$ standard deviation with 95\% confidence intervals for optimal trajectory experiments.

\begin{table*}[h]
\centering
\small
\begin{tabular}{llcccc}
    \toprule
    \textbf{Goals} & \textbf{Obs} & \textbf{GRAQL} & \textbf{BC-GRAIL} & \textbf{GAIL-GRAIL} & \textbf{AIRL-GRAIL} \\
    \midrule
    2 & 10\% & 0.80±0.31 & 0.73±0.33 & 0.71±0.09 & 0.66±0.11 \\
      &      & [0.56, 0.94] & [0.49, 0.90] & [0.64, 0.77] & [0.58, 0.74] \\
    2 & 20\% & 0.93±0.15 & 0.87±0.23 & 0.92±0.04 & 0.90±0.08 \\
      &      & [0.81, 0.99] & [0.68, 0.98] & [0.89, 0.95] & [0.84, 0.96] \\
    2 & 30\% & 1.00±0.00 & 0.93±0.20 & 0.96±0.02 & 0.94±0.06 \\
      &      & [1.00, 1.00] & [0.76, 1.00] & [0.95, 0.98] & [0.90, 0.98] \\
    4 & 10\% & 0.72±0.28 & 0.63±0.25 & 0.64±0.10 & 0.60±0.09 \\
      &      & [0.51, 0.84] & [0.45, 0.78] & [0.57, 0.71] & [0.54, 0.67] \\
    4 & 20\% & 0.90±0.18 & 0.82±0.22 & 0.89±0.04 & 0.85±0.07 \\
      &      & [0.75, 0.98] & [0.65, 0.91] & [0.86, 0.92] & [0.79, 0.90] \\
    4 & 30\% & 0.97±0.10 & 0.93±0.13 & 0.95±0.03 & 0.93±0.06 \\
      &      & [0.88, 0.99] & [0.82, 1.00] & [0.93, 0.97] & [0.89, 0.98] \\
    6 & 10\% & 0.51±0.16 & 0.37±0.18 & 0.58±0.09 & 0.53±0.08 \\
      &      & [0.39, 0.63] & [0.23, 0.51] & [0.51, 0.65] & [0.47, 0.59] \\
    6 & 20\% & 0.81±0.20 & 0.71±0.21 & 0.85±0.05 & 0.81±0.07 \\
      &      & [0.66, 0.95] & [0.55, 0.87] & [0.81, 0.88] & [0.76, 0.87] \\
    6 & 30\% & 0.87±0.14 & 0.79±0.13 & 0.92±0.03 & 0.89±0.07 \\
      &      & [0.76, 0.95] & [0.69, 0.88] & [0.90, 0.94] & [0.84, 0.94] \\
    \bottomrule
\end{tabular}
\caption{Detailed F$_1$-score statistics for optimal trajectories. Each cell shows mean±std in the first row and [95\% CI lower, upper] in the second row.}
\label{tab:optimal_f1_detailed}
\end{table*}

\section{Licenses and External Packages}
\label{sec:licenses}

Our experimental framework relies on several open-source environments and libraries. For transparency and reproducibility, we provide details regarding package versions, licenses, and citations below.

\paragraph{MiniGrid}
We used version 2.5.0 of the \texttt{minigrid} package~\citep{MinigridMiniworld23}, licensed under the \textbf{Apache License 2.0}. This license permits free use, modification, and distribution, subject to compliance with its terms.

\paragraph{Panda-Gym}
The Panda-Gym environment~\citep{gallouedec2021pandagym} was used via version 3.0.7 of the \texttt{panda-gym} package, distributed under the \textbf{MIT License}, which allows flexible use and redistribution.

\paragraph{DRACO and GRAQL Implementations}
The DRACO~\citep{nageris2024goal} and GRAQL~\citep{amado2022goal} baselines were either re-implemented according to the original papers or executed using codebases provided by the respective authors. Where applicable, we obtained explicit permission to use these implementations, ensuring accuracy and comparability. As in the DRACO implementation, we utilized PPO and SAC algorithms as provided by their reference implementations.

\paragraph{General}
All tools and environments used are either licensed for academic use or faithfully reproduced from cited research. This ensures full reproducibility of the GRAIL framework and all baselines evaluated in our experiments.

\end{document}